\documentclass[11pt]{article}

\usepackage[final]{acl}      % camera-ready 版
\usepackage{times}
\usepackage{latexsym}

\usepackage[T1]{fontenc}
\usepackage[utf8]{inputenc}

\usepackage{microtype}

\usepackage{inconsolata}

\usepackage{graphicx}

\usepackage{amsmath}
\usepackage{algorithm}
\usepackage{algorithmic}
\usepackage{subcaption}
\usepackage{enumitem}
\usepackage{amsmath}
\usepackage{amsfonts}
\usepackage{booktabs}
\usepackage{xspace}
\usepackage{multirow}
\usepackage{bm}
\usepackage{xcolor}
\usepackage{soul}
\usepackage{colortbl}
\usepackage{pifont}
\usepackage{makecell}
\usepackage{amssymb}

\usepackage[most]{tcolorbox}
\usepackage{listings}
\usepackage{xcolor}
\tcbset{
  colback=gray!2, colframe=gray!40,
  boxsep=0.6mm, left=1mm, right=1mm, top=0.6mm, bottom=0.6mm,
  arc=1mm
}
\lstdefinestyle{prompt}{
  basicstyle=\ttfamily\footnotesize,
  breaklines=true, columns=fullflexible,
  showstringspaces=false, keepspaces=true, tabsize=2
}
\newtcolorbox{promptbox}[2][]{enhanced, breakable=false, % 关键：禁用分段
  title={#2}, fonttitle=\bfseries, #1}

\title{The Art of Socratic Inquiry: A Framework for Proactive Template-Guided Therapeutic Conversation Generation}
\author{
  Mingwen Zhang$^{1}$, Minqiang Yang$^{1}$, Changsheng Ma$^{1}$, Yang Yu$^{1}$
  \AND 
  Hui Bai$^{1}$, Chen Xu$^{2}$, Xiangzhen Kong$^{1}$, Bin Hu$^{1}$ \\
  $^{1}$School of Information Science and Engineering, Lanzhou University \\
  $^{2}$School of Medical Technology, Beijing Institute of Technology \\
}

\begin{document}
\maketitle
\begin{abstract}
% Proactive questioning is fundamental in cognitive behavioral therapy (CBT), as it elicits latent concerns and facilitates guided discovery toward behavioral change.
% Yet, current psychological Large language models (LLMs) remain reactive, lacking mechanisms for timing questions and selecting CBT strategies. 
% In contrast, we propose the Socratic Inquiry Framework (SIF), a lightweight plug-and-play planner that augments pre-trained model to generate proactive, template-guided therapeutic conversations.
% Complementing SIF, we curate Socratic-QA, a CBT-aligned dataset that provides strategy-aware reasoning signals and enhances proactive questioning. 
% Integrating SIF with Socratic-QA yields a unified framework that substantially improves psychological LLMs in delivering personalized and context-aware therapeutic support.
% Experiments show that SIF integration improves proactive questioning, conversation depth, and response quality compared to baseline models. This work also outlines optimization opportunities for therapeutic dialogue systems and offers guidance for future research on proactive questioning in psychological LLMs.
Proactive questioning, where therapists deliberately initiate structured, cognition-guiding inquiries, is a cornerstone of cognitive behavioral therapy (CBT). 
Yet, current psychological large language models (LLMs) remain overwhelmingly reactive, defaulting to empathetic but superficial responses that fail to surface latent beliefs or guide behavioral change. To bridge this gap, we propose the \textbf{Socratic Inquiry Framework (SIF)}, a lightweight, plug-and-play therapeutic intent planner that transforms LLMs from passive listeners into active cognitive guides. SIF decouples \textbf{when to ask} (via Strategy Anchoring) from \textbf{what to ask} (via Template Retrieval), enabling context-aware, theory-grounded questioning without end-to-end retraining. Complementing SIF, we introduce \textbf{Socratic-QA}, a high-quality dataset of strategy-aligned Socratic sequences that provides explicit supervision for proactive reasoning. Experiments show that SIF significantly enhances proactive questioning frequency, conversational depth, and therapeutic alignment, marking a clear shift from reactive comfort to proactive exploration. Our work establishes a new paradigm for psychologically informed LLMs: not just to respond, but to guide. 
\end{abstract}

% \begin{figure*}[th]
%     \centering
%     % \includegraphics[width=\linewidth]{fig/dialogue_comparison2.png}
%     \includegraphics[width=\linewidth]{fig/dialog_comparison3.png}
%     \caption{Example dialogue showing how SIF elicit deeper reflection through structured questioning, compared to the shallow, declarative responses of generic PsyLLM.}
%     \label{fig:dialogue_comparison}
% \end{figure*}

\begin{figure*}[th]
    \centering
    \includegraphics[width=\linewidth]{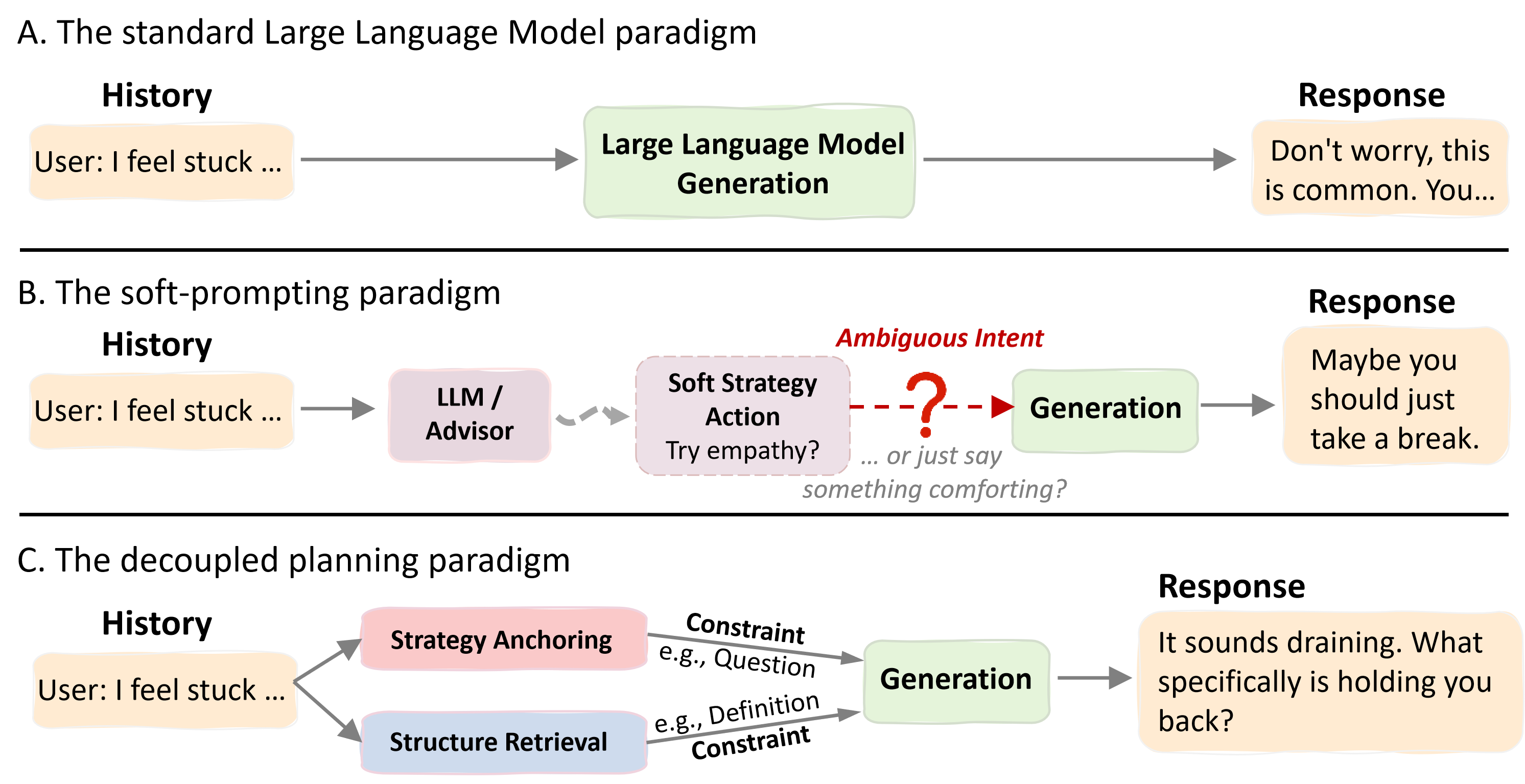}
    \caption{Comparison of dialogue control paradigms.
(a) The standard Large Language Model paradigm: Operates reactively based on raw probabilities.
(b) The soft-prompting paradigm: Relies on post-hoc rationale.
(c) The decoupled planning paradigm (Ours): Introduces proactive planning. }
    \label{fig:dialogue_comparison}
\end{figure*}

% \begin{figure*}[th]
%     \centering
%     % \includegraphics[width=\linewidth]{fig/dialogue_comparison2.png}
%     \includegraphics[width=\linewidth]{fig/dialog_comparison5.png}
%     \caption{Comparison of dialogue control paradigms.
% (a) The standard Large Language Model paradigm: Operates reactively based on raw probabilities.
% (b) The soft-prompting paradigm: Relies on post-hoc rationale.
% (c) The decoupled planning paradigm (Ours): Introduces proactive planning. }
%     \label{fig:dialogue_comparison}
% \end{figure*}

\section{Introduction}
Mental health disorders are a major global health challenge, with access to specialized care remaining limited due to a persistent shortage of trained clinicians, inequitable resource allocation, and high treatment costs \cite{1busch2024behind,2hall2023mental,3patel2018lancet,4ophuis2017cost}. 
Large language models (LLMs) have emerged as a promising avenue for scalable psychological support, delivering empathetic, open-ended conversations that simulate aspects of psychological support \cite{9trosper2009emotion,10garber2016developmental}. Recent efforts have further explored integrating LLMs into cognitive behavioral therapy (CBT), inspired systems through fine-tuning, prompt engineering, or memory-augmented architectures \cite{13raile2024usefulness,14stade2024large,15lee2024cactus,16wang2024patient,17na2024cbt}, reflecting a growing trend toward leveraging LLMs to address the mental health care problem.

As illustrated in Figure \ref{fig:dialogue_comparison}, existing LLMs remain overwhelmingly \textbf{reactive}: they mirror users’ emotions with comfort statements but rarely initiate the kind of structured, cognition-driven inquiry that defines effective intervention \cite{64qi-etal-2023-art}. While adapter~\cite{78hu-etal-2025-psyadvisor} has taken initial steps toward equipping psychological LLMs with proactive questioning capabilities, its strategy selection is grounded in post-hoc rationale, which apply LLMs to generate explanations that rationalize already-chosen strategies, rather than drive the decision-making process itself.

We argue this limitation stems from the absence of a mechanism to plan therapeutic intent. In clinical practice, therapists ask purposeful questions in over 64\% of utterances \cite{52Psy-Insight}, deliberately guiding clients to uncover hidden beliefs and challenge distortions. Current LLMs, however, entangle strategic reasoning with response generation, making it difficult to ensure consistent, theory-grounded questioning. Without explicit signals for \textbf{when} to question and \textbf{what} kind of question aligns with CBT goals, models tend to default to surface-level reassurance \cite{13raile2024usefulness,14stade2024large}.

To address this, we propose the \textbf{Socratic Inquiry Framework (SIF)}---a lightweight, plug-and-play therapeutic intent planner that equips any LLM with the ability to ask purposeful, CBT-aligned questions. At its core, SIF introduces a decoupled planning architecture composed of two synergistic modules: Strategy Anchoring, which determines what kind of therapeutic intent should guide the current turn, and Template Retrieval, which selects a Socratic questioning pattern aligned with that intent.
These signals jointly condition a conversation generator to produce natural, context-aware responses, enabling proactive, theory-informed dialogue without end-to-end retraining. To train this planner, we curate \textbf{Socratic-QA}, a high-quality dataset of strategy-aligned question sequences, filtered through a multi-dimensional clinical rubric that prioritizes therapeutic relevance over linguistic fluency. When integrated, SIF and Socratic-QA form a unified system that \textbf{shifts the dialogue paradigm from reactive comfort to proactive exploration}.
Our contributions are threefold:
\begin{itemize}
\item We reframe the challenge of therapeutic dialogue as one of proactive intent planning, not just empathetic response generation.
\item We introduce SIF, a modular framework that equips LLMs with clinically grounded, template-guided questioning capabilities through a decoupled architecture.
\item We demonstrate---via automatic metrics, human evaluation, and case studies---that SIF significantly improves proactive questioning frequency, conversational depth, and therapeutic alignment, moving psychological LLMs closer to the role of an active guide rather than a passive companion. 
\end{itemize}

% % ------------------------------------------------------------------------
\begin{figure*}[htbp]
    \centering
    \includegraphics[width=\linewidth]{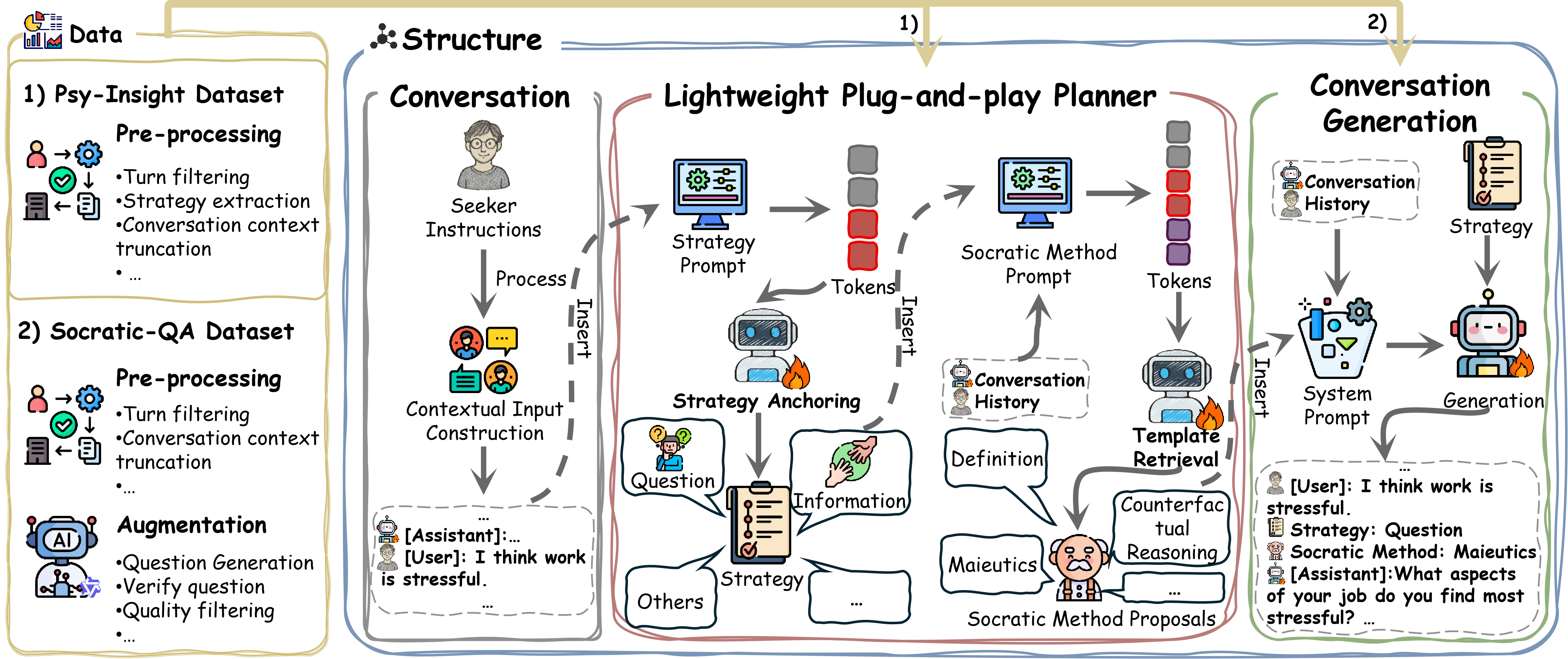}
    \caption{The proposed framework SIF integrates LPP module with CG module. Within LPP, the SA component first extracts conversational policies from seeker utterances; these policies then guide the TR module to select suitable question templates. The combined outputs of SA and TR condition the CG component, which is fine-tuned on our domain-specific Socratic-QA dataset to produce context-aware, goal-directed responses.}
    \label{fig:structure}
\end{figure*}
% % ------------------------------------------------------------------------

\section{Related Work}
\label{sec:related}

\subsection{LLMs in CBT}
LLMs~\cite{74xu2024small,75zhang2024cpsycoun,76zhang2023ask} have been increasingly employed in CBT, particularly for tasks such as motivational interviewing and detecting cognitive distortions, to support individuals with affective disorders~\cite{34lai2023psy, 35do2022pair}.
To enhance conversational quality, recent approaches incorporate prompt tuning and memory-augmented techniques~\cite{37ding2022improving, 38maddela2023training}, yielding better adherence with CBT principles than earlier rule-based systems.
To mitigate the scarcity of domain-specific training data, resources such as the ExTES dataset~\cite{32zheng2023building} and the SMILE framework~\cite{33qiu2023smile} have been introduced to synthesize realistic therapeutic conversations. 
Furthermore, specialized models including CBT-LLM~\cite{17na2024cbt} and CoCoA~\cite{43lee2024cocoa} integrate personalized prompting and memory mechanisms to improve contextual relevance and seeker utterances engagement.
Prior CBT-LLMs treat therapeutic dialogue as an end-to-end generation task, entangling strategic intent with surface-level response generation, resulting in reactive and  generic interactions. SIF decouples therapeutic intent from response, enabling proactive, theory-aligned questioning. 

\subsection{Socratic Questioning}
Socratic Questioning is a structured, inquiry-driven technique designed to promote critical reflection on personal beliefs and underlying assumptions~\cite{46paul2014critical,47yang2005using}. 
Recent efforts have extended this method to LLMs, with promising yet limited success. 
For instance, CPsyCoun implements multi-turn Socratic questioning through a four-step counseling model that encourages seeker utterances to articulate, examine, and revise their personal concerns~\cite{19zhang2024cpsycoun}. 
The SPL framework employs an iterative prompting mechanism that refines seeker utterance responses through repeated cycles of reasoning and feedback, promoting deeper cognitive engagement~\cite{50zhang2024spl}. SocREval~\cite{69he2024socreval} applies Socratic prompts within a structured evaluation pipeline, guiding LLMs through evidence appraisal, hypothesis critique, and reflective analysis. 
Additionally, a generalized Socratic Method framework~\cite{51chang2023prompting} explicitly decomposes the reasoning process into four stages: claim formulation, evidence evaluation, counterargument analysis, and exploration of alternatives. 
Existing Socratic methods rely on static prompting or end-to-end fine-tuning, lacking explicit intent modeling or timing control.
SIF operationalizes Socratic methods as executable planning signals, ensuring intentional and context-aware inquiry. 
These frameworks demonstrate that LLMs can benefit from structured reasoning scaffolds modeled after classical Socratic techniques.

\section{Socratic Inquiry Framework}
\label{sec:method}
\subsection{Overview}
SIF enhances proactive reasoning in CBT conversations through a Lightweight Plug-and-play Planner (LPP) that steers a pre-trained Conversation Generation (CG) model. As illustrated in Figure~\ref{fig:structure}, the LPP first determines the appropriate therapeutic strategy and Socratic questioning method from the dialogue context, then conditions the CG to produce a response that is both context-aware and methodologically grounded. 
Crucially, LPP is trained on strategy annotations from the Psy-Insight dataset and CG is trained on \textbf{Socratic-QA}, a high-quality corpus of Socratic method-aligned sequences, providing explicit supervision for \textbf{when} to question and \textbf{what} kind of question to ask. This decoupled architecture, grounded in clinically curated data, ensures that every utterance is guided by explicit therapeutic intent, transforming the LLM from a passive responder into an active cognitive guide.

\subsection{Task Definition}
% Given a multi-turn therapeutic conversation $C$:
% \begin{equation}
% C = \{c_1, \ldots, c_{|C|}\},
% \end{equation}
% with each turn $c_i = (x_i, y_i)$, and a set of CBT techniques $T = \{t_1, \ldots, t_{|T|}\}$, where each $t_j \in T$ denotes a structured therapeutic strategy, the goal is to select an appropriate technique $t_j$ for a seeker utterance $x_i$ and generate a response $y_i$ that applies $t_j$ to support therapeutic reasoning while preserving contextual and semantic integrity.
Given a multi-turn therapeutic conversation context $C = \{c_1, \dots, c_{|C|}\}$, where each turn $c_i = (x_i, y_i)$ consists of a seeker utterance $x_i$ and a supporter response $y_i$, the task is to select a CBT-aligned therapeutic technique $t_j$ from a predefined set $\mathcal{T} = \{t_1, \dots, t_{|\mathcal{T}|}\}$ and generate a response $y_i$ that applies $t_j$ to advance therapeutic reasoning, subject to the constraints of contextual coherence and semantic integrity. Formally, the objective is to maximize the alignment between the generated response and the intended therapeutic function, while ensuring the output remains grounded in the dialogue history and adheres to clinical norms of Socratic inquiry.

\subsection{Lightweight Plug-and-play Planner}
The LPP serves as the cognitive core of SIF, responsible for translating raw dialogue context into structured therapeutic intent. It operates in two sequential stages, Strategy Anchoring (SA) and Template Retrieval (TR), each designed to mirror a distinct layer of clinical reasoning observed in expert CBT practice. This two-stage decomposition not only aligns with the hierarchical nature of therapeutic decision-making but also enables modular training and plug-and-play deployment without modifying the underlying language model.

\subsubsection{Strategy Anchoring}
Given a multi-turn conversation context $C = \{c_1, \dots, c_{|C|}\}$, where each turn $c_i = (x_i, y_i)$ consists of a seeker utterance $x_i$ and a historical supporter response $y_i$, the SA module predicts a high-level supportive strategy $s \in \mathcal{S}$ for the current turn. The strategy set $\mathcal{S}$ comprises ten clinically validated categories: $S=$ \textit{\{question, reflection of feelings,
self-disclosure, others, information,
providing suggestions, role-play,
restatement or paraphrasing,
unknown, affirmation and reassurance\}.}
This taxonomy is derived from the Psy-Insight dataset \cite{52Psy-Insight}, we show the distribution statistics of strategies and more details in Appendix \ref{sec:Psy-Insight strategy}.

To handle long dialogue histories while preserving computational efficiency, we left-truncate the context to $\hat{C} = \{c_k, \ldots, c_{|C|}\}$, retaining only the most recent turns that fit within the model’s token budget. An instruct-tuned encoder $f_{\theta_1}$~\cite{55bai2023qwen} processes the concatenated sequence $(\hat{C}, x_i)$ and produces a contextualized hidden representation $h_i$ for the current seeker utterance:
\begin{equation}
    h_i = f_{\theta_1}(\hat{C}, x_i).
\end{equation}
A linear projection layer then maps this representation to a logits vector over the strategy space:
\begin{equation}
    z_i = W h_i + b, \quad W \in \mathbb{R}^{|\mathcal{S}| \times d},
\end{equation}
where $d$ is the hidden dimension. The resulting probability distribution over strategies is obtained via softmax:
\begin{equation}
    p(s \mid h_i) = \mathrm{softmax}(z_i).
\end{equation}
At inference time, the predicted strategy is selected as the argmax:
\begin{equation}
    \hat{s} = \arg\max_{s \in \mathcal{S}} p(s \mid h_i).
\end{equation}
This discrete strategy label $\hat{s}$ functions as a high-level intent anchor, constraining the subsequent generation to fulfill a specific therapeutic role, ensuring that a turn labeled \textit{question} indeed initiates an inquiry rather than offering reassurance.

\subsubsection{Template Retrieval}

Once the macro-strategy $\hat{s}$ is determined, the TR module refines the plan into a concrete Socratic questioning pattern. While SA decides \textbf{what kind of support to provide}, TR specifies \textbf{how to formulate the question} in a manner consistent with classical Socratic pedagogy. The template space $\mathcal{T}$ includes six methodologically distinct categories: \textit{definition}, \textit{counter-questioning}, \textit{maieutics}, \textit{dialectics}, \textit{counterfactual reasoning}, and \textit{other}. Each category corresponds to a well-established cognitive operation, and detailed descriptions are given in Appendix~\ref{sec:TR model}).

TR is implemented as a fine-tuned classifier that takes the same truncated context $\hat{C}$ and seeker utterance $x_i$ as input, but conditions its prediction on the strategy $\hat{s}$ implicitly through training data alignment. Using a second instruct encoder $f_{\theta_2}$, it produces a 6-dimensional logits vector:
\begin{equation}
    z_i = f_{\theta_2}(\hat{C}, x_i) \in \mathbb{R}^6,
\end{equation}
where each dimension corresponds to a Socratic method in $\mathcal{T}$. The class probabilities are computed as:
\begin{equation}
    p_{i,c} = \frac{\exp(z_{i,c})}{\sum_{k=1}^{6} \exp(z_{i,k})}, \quad c \in \{1,\dots,6\}.
\end{equation}
Given ground-truth annotations from the Socratic-QA dataset, the model is trained with cross-entropy loss. At inference, the selected template is:
\begin{equation}
    \hat{t} = \arg\max_{c \in \{1,\dots,6\}} p_{i,c}, \quad \hat{t} \in \mathcal{T}.
\end{equation}
The pair $(\hat{s}, \hat{t})$ thus constitutes a compact yet expressive planning signal that captures both the dialogic function and cognitive structure of the intended response.

\subsection{Conversation Generation}
The CG module realizes the planned intent in natural language. To ensure that therapeutic reasoning is explicitly grounded in the generation process, CG conditions on the full planning signal $(\hat{s}, \hat{t})$ alongside the dialogue history $C$ and the latest seeker utterance $x_i$. The input sequence is constructed through deterministic token-level concatenation:
\begin{equation}
    \mathrm{seq} = \mathrm{Sequence}(\hat{s}, \hat{t}, C, x_i),
\end{equation}
where $\mathrm{Sequence}(\cdot)$ denotes an ordered composition that preserves the semantic roles of each component. This design ensures that high-level intent is injected before the generative process begins, rather than emerging stochastically from language modeling alone.

Subsequently, a small-scale LLM was employed as the backbone to balance the therapeutic reasoning fidelity and computational efficiency, fine-tuned via LoRA~\cite{54hu2022lora} to adapt to the CBT domain while preserving general language capabilities as:
\begin{equation}
    y_i = \mathrm{LLM}_{\theta}^{\mathrm{LoRA}}(\mathrm{seq}).
\end{equation}
During generation, the strategy $\hat{s}$ acts as a functional constraint enforcing interrogative syntax, while the Socratic method $\hat{t}$ shapes the internal reasoning pattern. This dual conditioning enables the model to produce responses that are simultaneously fluent, contextually grounded, and methodologically rigorous.

\subsection{Dataset Construction}
\label{subsec:dataset}
The Socratic-QA dataset is constructed through a two-stage pipeline designed to produce high-fidelity, strategy-aligned Socratic questioning sequences suitable for training therapeutic intent planners. This process begins with automatic question generation and concludes with rigorous quality-based filtering, ensuring that the final corpus prioritizes counseling relevance, emotional attunement, and methodological consistency.

\subsubsection{Question Generation}
Initial question candidates are derived from multi-turn dialogues in the EmoLLM dataset~\cite{53yang2024emollm}. For each seeker utterance, we first extract the surrounding conversational context to preserve pragmatic grounding. Domain-specific Socratic templates are then applied to generate context-aware follow-up questions. The generation process explicitly enforces three criteria: (1) Openness, to encourage elaboration rather than yes/no responses. (2) Emotional resonance, to maintain empathetic alignment with the seeker’s affective state. (3) Non-directiveness, to avoid prescriptive phrasing such as “you should” or “try this.” This yields an initial pool of fluent, context-sensitive question candidates.

\subsubsection{Quality Filtering}
To distill high-quality samples from this pool, we employ a multi-dimensional assessment framework that evaluates each candidate along seven clinically informed dimensions: guidance (20\%), empathy (20\%), semantic relevance (15\%), interrogative structure (15\%), conciseness (10\%), diversity (10\%), and tone friendliness (10\%). All scores are normalized to $[0,1]$, and semantic relevance is quantified via cosine similarity between the candidate and the seeker utterance, augmented with domain-specific keyword matching to capture professional consultation patterns. Notably, for anxiety-related dialogues, the empathy weight is dynamically increased to prioritize emotionally supportive phrasing.

To operationalize this rubric, we implement a contrastive preference mechanism as outlined in Appendix Algorithm~\ref{alg:quality_filtering}. For each context, two candidate questions are generated and scored independently. The higher-scoring candidate is labeled as \textit{chosen}, while the lower-scoring one is marked as \textit{rejected}; ties result in one being retained as \textit{chosen} and the other discarded. This pairwise comparison ensures that only responses demonstrating clear superiority in therapeutic quality are retained. As a result, 75.7\% of initial candidates are filtered out, yielding a final curated set of 17,981 high-quality samples (see Appendix \ref{sec: Distribution of Samples}). This stringent filtering guarantees that Socratic-QA provides precise supervision for learning \textbf{when} to question and \textbf{how} to structure that question in alignment with clinical best practices.

% % -------------------------基于单轮提问提示CBT-specific--------------------------------
\begin{table*}[ht]
\begin{center}
\scalebox{0.70}{%
\begin{tabular}{cccccccc}
\toprule
\textbf{Type} & \textbf{Method} & \textbf{Bert Score$\uparrow$} & \textbf{Bleurt$\uparrow$} & \textbf{Distinct rate$\uparrow$} & \textbf{Chrf$\uparrow$} & \textbf{Meteor$\uparrow$} & \textbf{Rouge$\uparrow$} \\
\midrule
\multirow{4}{*}{\parbox{2cm}{\centering Zero-shot\\Large-scale\\General LLMs}}
 & \cellcolor{gray!20} Deepseek R1 671B (Sig.) & \cellcolor{gray!20} 0.69 (\textbf{+0.08}) & \cellcolor{gray!20} 0.45 (\textbf{+0.14}) & \cellcolor{gray!20} 0.57 ({-0.17}) & \cellcolor{gray!20} 0.06 (\textbf{+0.03}) & \cellcolor{gray!20} 0.35 (\textbf{+0.10}) & \cellcolor{gray!20} 0.26 (\textbf{+0.12}) \\
 & \cellcolor{gray!20} Deepseek R1 671B (Mul.) & \cellcolor{gray!20} 0.89 (\textbf{+0.01}) & \cellcolor{gray!20} 0.38 (\textbf{+0.01}) & \cellcolor{gray!20} 0.56 (\textbf{+0.02}) & \cellcolor{gray!20} 0.66 (\textbf{+0.01}) & \cellcolor{gray!20} 0.33 (\textbf{+0.04}) & \cellcolor{gray!20} 0.41 (\textbf{+0.07}) \\
 & \cellcolor{gray!20} Deepseek V3 671B (Sig.) & \cellcolor{gray!20} 0.69 (\textbf{+0.06}) & \cellcolor{gray!20} 0.32 (\textbf{+0.12}) & \cellcolor{gray!20} 0.63 ({-0.04}) & \cellcolor{gray!20} 0.08 (\textbf{+0.03}) & \cellcolor{gray!20} 0.33 (\textbf{+0.12}) & \cellcolor{gray!20} 0.27 (\textbf{+0.13}) \\
 & \cellcolor{gray!20} Deepseek V3 671B (Mul.) & \cellcolor{gray!20} 0.91 (\textbf{+0.02}) & \cellcolor{gray!20} 0.41 (\textbf{+0.00}) & \cellcolor{gray!20} 0.52 (\textbf{+0.05}) & \cellcolor{gray!20} 0.70 (\textbf{+0.03}) & \cellcolor{gray!20} 0.39 (\textbf{+0.07}) & \cellcolor{gray!20} 0.47 (\textbf{+0.07}) \\
\midrule

\multirow{6}{*}{\parbox{2cm}{\centering Zero-shot\\Psychological\\LLMs}}
 & SoulChat 7B (Sig.)    & 0.64 (-0.01) & -0.10 (+0.00) & 0.55 (\textbf{+0.01}) & 0.06 (+0.00) & 0.15 (+0.00) & 0.14 (-0.01) \\
 & SoulChat 7B (Mul.)    & 0.90 (+0.00) & 0.41 (+0.00) & 0.25 (-0.03) & 0.65 (+0.00) & 0.25 (-0.02) & 0.39 (\textbf{+0.01}) \\
 & CBT-LLM 7B (Sig.)     & 0.63 (\textbf{+0.04}) & -1.04 (+0.00) & 0.27 (-0.04) & 0.07 (-0.02) & 0.04 (+0.00) & 0.10 (\textbf{+0.02}) \\
 & CBT-LLM 7B (Mul.)     & 0.89 (+0.00) & \textbf{0.41 (+0.02)} & 0.12 (+0.00) & 0.58 (\textbf{+0.01}) & 0.13 (\textbf{+0.01}) & 0.26 (\textbf{+0.02}) \\
 & CBT-BENCH 7B (Sig.)   & 0.65 (\textbf{+0.02}) & -0.28 (\textbf{+0.32}) & \textbf{0.59 (+0.03)} & 0.08 (+0.00) & 0.16 (\textbf{+0.08}) & 0.18 (\textbf{+0.04}) \\
 & CBT-BENCH 7B (Mul.)   & \textbf{0.91 (+0.02)} & \underline{0.40 (\textbf{+0.01})} & 0.27 (+0.00) & 0.65 (\textbf{+0.2}) & 0.19 (\textbf{+0.01}) & 0.31 (\textbf{+0.02}) \\
\midrule

\multirow{8}{*}{\parbox{2.8cm}{\centering Fine-tuned\\Small-scale\\Open-source\\LLMs}}
 & Llama 7B (Sig.)    & 0.62 (+0.00) & -0.67 (\textbf{+0.10}) & 0.36 (-0.02) & 0.06 (+0.00) & 0.08 (\textbf{+0.02}) & 0.13 (\textbf{+0.05}) \\
 & Llama 7B (Mul.)    & \underline{0.90 (\textbf{+0.02})} & \textbf{0.41 (+0.00)} & 0.32 (-0.02) & 0.66 (\textbf{+0.03}) & 0.27 (\textbf{+0.09}) & \underline{0.46 (\textbf{+0.16})} \\
 & Qwen2 7B (Sig.)    & \underline{0.67 (\textbf{+0.05})} & \underline{0.21 (\textbf{+0.88})} & 0.56 (\textbf{+0.12}) & \underline{0.08 (\textbf{+0.03})} & \underline{0.27 (\textbf{+0.20})} & \underline{0.23 (\textbf{+0.14})} \\
 & Qwen2 7B (Mul.)    & \underline{0.90 (\textbf{+0.02})} & \textbf{0.41 (+0.00)} & \underline{0.46 (\textbf{+0.13})} & \underline{0.67 (\textbf{+0.06})} & \underline{0.30 (\textbf{+0.16})} & \underline{0.41 (\textbf{+0.16})} \\
 & Chatglm 7B (Sig.)  & 0.65 (\textbf{+0.02}) & -0.32 (\textbf{+0.17}) & 0.48 (\textbf{+0.03}) & \underline{0.08 (\textbf{+0.02})} & 0.13 (\textbf{+0.04}) & 0.17 (\textbf{+0.05}) \\
 & Chatglm 7B (Mul.)  & 0.89 (\textbf{+0.01}) & \underline{0.40 (-0.01)} & 0.45 (\textbf{+0.10}) & 0.65 (\textbf{+0.02}) & 0.25 (\textbf{+0.06}) & 0.35 (\textbf{+0.05}) \\
 & Alpaca 7B (Sig.)   & 0.58 (+0.00) & -0.91 (\textbf{+0.03}) & 0.25 (\textbf{+0.01}) & 0.05 (\textbf{+0.01}) & 0.05 (\textbf{+0.01}) & 0.12 (\textbf{+0.01}) \\
 & Alpaca 7B (Mul.)   & \textbf{0.91 (+0.02)} & 0.41 (\textbf{+0.04}) & 0.27 (+0.00) & 0.61 (\textbf{+0.15}) & 0.29 (\textbf{+0.12}) & \textbf{0.48 (+0.13)} \\
\midrule

\multirow{2}{*}{\parbox{2cm}{\centering Ours}}
 & SIF 7B (Sig.)        & \textbf{0.68 (+0.06)} & \textbf{0.28 (+0.20)} & \underline{0.58 (\textbf{+0.06})} & \textbf{0.08 (+0.04)} & \textbf{0.30 (+0.13)} & \textbf{0.25 (+0.14)} \\
 & SIF 7B (Mul.)        & \textbf{0.91 (+0.05)} & \textbf{0.41 (+0.00)} & \textbf{0.47 (+0.07)} & \textbf{0.69 (+0.02)} & \textbf{0.32 (+0.02)} & \underline{0.41 (\textbf{+0.02})} \\
\bottomrule
\end{tabular}
}
\caption{Evaluation results of CBT conversation generation under different settings. Values in parentheses indicate the performance gains when integrating our \textbf{LPP}. \textbf{Sig.} refers to single-utterance input, \textbf{Mul.} to multi-turn interaction. 
Large-scale General-purpose LLMs are shown for reference but excluded from best/second-best highlighting.
}
\label{tab:main_result} 
\end{center}
\end{table*}
% %------------------------------------------------------

\section{Experiments}
\subsection{Experimental Settings}

\paragraph{Baselines and Evaluation Protocol.} 
We evaluate the generalizability and plug-and-play capability of SIF by integrating its LPP with a diverse set of backbone language models. Specifically, for each baseline model, we compare two settings: (1) the model’s native zero-shot or fine-tuned output, and (2) the same model used as the CG component conditioned on LPP's planning signals. The performance gain quantifies the contribution of proactive intent planning across architectures. Full implementations are detailed in Appendix~\ref{sec:Implementation}.

Our baseline suite includes:  
(1) Large-scale general LLMs (>30B parameters): including Deepseek R1 671B~\cite{58guo2025deepseek}, Deepseek V3 671B~\cite{59liu2024deepseek}, evaluated in zero-shot mode.  
(2) Psychological LLMs, including SoulChat~\cite{73chen2023soulchat},  CBT-LLM~\cite{71na2024cbt}, CBT-BENCH~\cite{72zhang2024cbt}, also in zero-shot mode.  
(3) Small-scale open-source LLMs (<10B parameters): Small-scale open-source LLMs (<10B), such as LLaMA 7B~\cite{61touvron2023llama}, Qwen2 7B, ChatGLM 7B~\cite{62glm2024chatglm}, and Alpaca 7B~\cite{63Chinese-LLaMA-Alpaca}, fine-tuned on the Socratic-QA training set to ensure fair comparison with SIF.  
All evaluations are conducted under both single-turn (Sig.) and multi-turn (Mul.) dialogue settings, using a unified turn-based protocol.

\paragraph{Dataset.}
We construct evaluation samples from the EmoLLM dataset~\cite{52Psy-Insight}, extracting multi-turn counseling sessions while preserving full conversational context. For each seeker utterance, we generate a Socratic follow-up question using the template-based pipeline described in Section~\ref{subsec:dataset}, resulting in a corpus of strategy-aligned therapeutic prompts.
Crucially, we perform session-level splitting, all turns from the same counseling session are assigned exclusively to either training or test sets, ensuring no overlap in conversation IDs across splits. This prevents data leakage and guarantees that models are evaluated on unseen dialogue trajectories. The final split yields 17,981 training samples (Socratic-QA) and 2,697 held-out test samples, all of which conform to professional counseling standards after multi-dimensional filtering.

\paragraph{Evaluation Metrics.} 
We evaluate from three aspects: Automatic evaluation metrics include BERTScore, BLEURT, ROUGE, METEOR, chrF, and Distinct-$n$ to assess semantic coherence, lexical overlap, and response diversity. Proactive Questioning Ability (PQA)~\cite{77shen2024large} value measures the degree to which a response initiates directive, cognition-driven questions aligned with CBT objectives.
Human evaluation metrics are conducted by nine mental health professionals (eight graduate students and one senior counselor) on four dimensions: Strategy Comprehensiveness (SC; 0--2), Professionalism (Prof; 0--3), Authenticity (Auth; 0--3), and Ethical Safety (ES; 0--1). Model identities are anonymized, and evaluators are blinded to the presence of LPP to minimize bias. Full rating protocols are detailed in Appendix~\ref{sec:Human Evaluation Process}.

% %----------------------------------自动测评--------------------------------
\begin{table*}[ht]
\begin{center}
\scalebox{0.75}{%
\begin{tabular}{cccccccc}
\toprule
\multirow{2}{*}{\textbf{Type}} & \multirow{2}{*}{\textbf{Method}}
& \multicolumn{4}{c}{\textbf{Human Evaluation Metrics}} & \multicolumn{2}{c}{\textbf{PQA}}\\
\cmidrule(lr){3-6}
\cmidrule(lr){7-8}
& &\textbf{SC.}$\uparrow$ &\textbf{Prof.}$\uparrow$ &\textbf{Auth.}$\uparrow$ &\textbf{ES.}$\uparrow$  &\textbf{Sig.}$\uparrow$ &\textbf{Mul.}$\uparrow$ \\
\midrule

\multirow{2}{*}{\parbox{4.2cm}{\centering Zero-shot Large-scale\\General LLMs}} 
 & \cellcolor{gray!20} Deepseek R1 671B & \cellcolor{gray!20} 1.90 (\textbf{+0.02}) & \cellcolor{gray!20} 2.84 (+0.00) & \cellcolor{gray!20} 2.78 (\textbf{+0.02}) & \cellcolor{gray!20} 1.00 & \cellcolor{gray!20} 0.99 (\textbf{+0.58}) & \cellcolor{gray!20} 0.95 (\textbf{+0.19}) \\
 & \cellcolor{gray!20} Deepseek V3 671B & \cellcolor{gray!20} 1.94 (\textbf{+0.06}) & \cellcolor{gray!20} 2.78 (\textbf{+0.06}) & \cellcolor{gray!20} 2.75 (+0.00) & \cellcolor{gray!20} 1.00 & \cellcolor{gray!20} 0.99 (\textbf{+0.68}) & \cellcolor{gray!20} 0.91 (\textbf{+0.38}) \\
 % & Qwq plus 32B & 1.83 & 2.79 & \textbf{2.91} & 1.00  & 0.99 (\textbf{+0.58}) & 0.93 (\textbf{+0.16}) \\
\midrule

\multirow{3}{*}{\parbox{2.8cm}{\centering Zero-shot\\Psychological\\LLMs}} 
 & SoulChat 7B & 1.75 (\textbf{+0.04}) & 2.43 (\textbf{+0.24}) & 2.57 (+0.00) & 1.00 & 0.19 (\textbf{+0.03}) & 0.16 (\textbf{+0.08})\\
 & CBT-LLM 7B & 1.74 (\textbf{+0.01}) & 2.52 (\textbf{+0.08}) & \underline{2.76 (\textbf{+0.01})} & 1.00 & 0.54 (\textbf{+0.06}) & 0.14 (\textbf{+0.01})\\
 & CBT-BENCH 7B & 1.82 (+0.00) & 2.24 (\textbf{+0.08}) & 2.32 (\textbf{+0.03}) & 1.00 & 0.98 (\textbf{+0.20}) & 0.80 (\textbf{+0.22})\\
\midrule

\multirow{4}{*}{\parbox{2.8cm}{\centering Fine-tuned\\Small-scale\\Open-source\\LLMs}} 
 & Llama 7B & \underline{1.83 (\textbf{+0.05})} & 2.50 (\textbf{+0.06}) & 2.59 (\textbf{+0.09}) & 1.00 & 0.44 ($-$0.07) & 0.48 (\textbf{+0.34}) \\
 & Qwen2 7B & 1.74 (\textbf{+0.04}) & 2.64 (\textbf{+0.14}) & 2.54 (\textbf{+0.06}) & 1.00 & \textbf{0.99} (\textbf{+0.86}) & 0.96 (\textbf{+0.87}) \\
 & Chatglm 7B & 1.74 (\textbf{+0.02}) & \underline{2.70 (\textbf{+0.02})} & \underline{2.76 (\textbf{+0.04})} & 1.00 & 0.62 (\textbf{+0.53}) & 0.76 (\textbf{+0.68}) \\
 & Alpaca 7B & 1.49 (\textbf{+0.02}) & 2.05 (\textbf{+0.10}) & 2.09 (\textbf{+0.17}) & 1.00 & 0.34 (\textbf{+0.06}) & 0.81 (\textbf{+0.08}) \\
\midrule

\multirow{1}{*}{\parbox{2.8cm}{\centering Ours}} 
 & SIF 7B & \textbf{1.90 (+0.02)} & \textbf{2.79 (+0.15)} & \textbf{2.82 (+0.13)} & \textbf{1.00} & \textbf{0.99} (\textbf{+0.46}) & \textbf{0.97} (\textbf{+0.41}) \\
\bottomrule
\end{tabular}
}
\caption{Human and PQA evaluations of conversation models, with LPP gains in PQA shown in parentheses.}
\label{tab:auto_eval} 
\end{center}
\end{table*}
% %-------------------------------------------------------------------

\subsection{Results and Evaluation}
\paragraph{Automatic evaluation metrics.}
Table~\ref{tab:main_result} reveals three key findings. 
First, integrating the LPP yields substantial and consistent gains across all zero-shot large-scale general-purpose LLMs. In single-turn settings, Deepseek R1 and Deepseek V3 achieve Rouge improvements of 0.12, 0.13 respectively, nearly doubling their baseline scores. This demonstrates that LPP effectively injects structured therapeutic intent into models that otherwise lack domain-specific guidance.

Second, small-scale open-source LLMs fine-tuned on Socratic-QA also benefit significantly, with Qwen2-7B gaining 0.88 in Bleurt  in single-turn mode. This powerfully confirms that LPP complements fine-tuning by providing explicit planning signals that enhance coherence and proactivity.
Third, specialized psychological LLMs show limited or even negative gains (e.g., CBT-LLM’s distinct rate drops by 0.04 in single-turn). We attribute this to a mismatch between LPP’s external strategy signals and the internal CBT knowledge already embedded in these models through prior training or prompting. When external planning constraints conflict with internal priors, performance may stagnate or degrade.

Critically, our full system, SIF (LPP + Qwen2.5-7B as the backbone of CG), achieves state-of-the-art performance among all 7B-scale models, outperforming baselines across nearly all automatic metrics. This confirms that the decoupled architecture of SIF is not only effective but also optimally balanced for the CBT task.

\paragraph{PQA value.}
As shown in Table~\ref{tab:auto_eval}, integrating the LPP yields substantial PQA gains across most backbones: 0.46 in single-turn  settings for SIF, with even larger improvements for certain models (e.g., Qwen2-7B: +0.87 in Multi-turn mode).

The magnitude and direction of these gains reveal three distinct patterns. First, zero-shot large-scale general-purpose LLMs exhibit consistent and significant PQA improvements. This indicates that, while these models possess strong linguistic fluency, they lack explicit therapeutic intent. While LPP effectively supplies the missing “when to ask” and “what to ask” signals, transforming generic empathy into goal-directed inquiry.

Second, small-scale open-source LLMs fine-tuned on Socratic-QA also benefit markedly. demonstrating that LPP complements data-driven adaptation by providing structured planning scaffolds that prevent fine-tuned models from regressing to passive response patterns.

Third, and most notably, specialized psychological LLMs show limited PQA gains. This may because when the planner’s strategy or template contradicts the model’s learned behavior, the generation process may become inconsistent, diluting the intended proactive effect. This suggests that for models already optimized for CBT, external planning modules must be carefully aligned with their internal knowledge representations to avoid interference.

\paragraph{Human Evaluation metrics}
Table~\ref{tab:auto_eval} presents a comprehensive comparison of SIF against a wide range of baselines under multi-turn dialogue settings. Notably, SIF achieves the highest Authenticity score among all models, surpassing even large-scale models with parameter counts exceeding 671B. This result is particularly striking given the vast disparity in scale, and strongly suggests that architectural design outweighs raw model size in delivering clinically authentic interactions.

Furthermore, when compared to other 7B-class open-source models SIF demonstrates consistent and significant advantages across all human-rated dimensions. These gains are directly attributable to SIF's decoupled planning-generation architecture: by explicitly separating therapeutic intent planning (via SA and TR) from language realization, SIF ensures that every response is grounded in a coherent clinical rationale, rather than relying on implicit patterns learned from data. This structured approach enables even a modest 7B backbone to produce responses that feel more empathetic, contextually attuned, and professionally sound than those generated by much larger models operating without explicit planning.

\subsection{Case Study}
To validate the effectiveness of SIF, we present a representative case in Figure~\ref{fig:example_conversation}, with additional examples discussed in Appendix~\ref{sec:Case Study Details}. 
The seeker, initially reporting anhedonia, is guided to recall childhood drawing, translate it into present agency (“doodle to engage manual and cognitive faculties”), and commit to a next step. Compared to a generic LLM, SIF yields denser rationales, higher turn-level coherence, and clearer markers of therapeutic traction. Mechanistically, separating “what to ask” from “how to ask” reduces drift and template-like repetition; questions move from surface validation to discrepancy exploration and guided discovery, tightening inter-turn constraints and enabling deeper cognitive work without added verbosity. Although this single case limits generality, it reveals baseline failure modes (premature reassurance, ungrounded advice) and shows how SIF counters them via evidence seeking, hypothesis testing, and graded commitment—producing clinician-traceable, plan-bearing insights.

\begin{figure}%[htbp]
    \centering
    \includegraphics[width=1.0\linewidth]{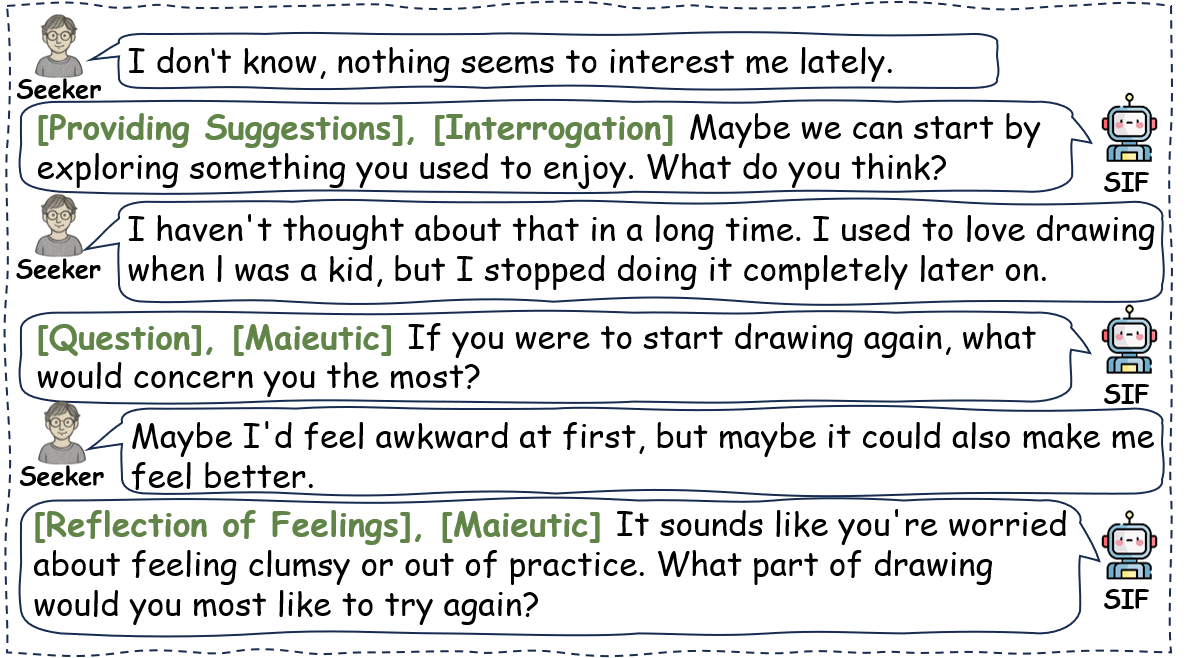}
    \caption{An example of conversation generation. All conversations include predictions from both the SA and TR modules.}
    \label{fig:example_conversation}
\end{figure}

%------------------------------ablation--------------------------------
\begin{table}%[htbp]
\centering
% \vspace{-0.5em}
\scalebox{1.0}{\begin{tabular}{ccccc}
\hline
\textbf{Model Variant} &\textbf{ SC.} & \textbf{Prof.} & \textbf{Auth.} & \textbf{PQA} \\
\hline
% baseline &1.72 &2.52 &2.58 &0.45 \\ 
w/o SA       & 1.82 & 2.52 & 2.68 & 0.94 \\
w/o TR    & 1.88 & 2.54 & 2.74 & 0.95 \\
w/o SocraticQA       & 1.84 & 2.40 & 2.64 & 0.85 \\
Ours        & \textbf{1.90} & \textbf{2.79} & \textbf{2.82} & \textbf{0.97}\\ 
\hline
\end{tabular}
}
\caption{Impact of  semantic behavioral alignment on CBT performance.}
\label{tab:ablation}
\end{table}
%-------------------------------------------------------------------

\subsection{Ablation Study}
Table~\ref{tab:ablation} reveals that removing Socratic QA induces the most severe performance degradation, substantiating this prevents generation from degenerating into shallow matching. Furthermore, the absence of Strategy Anchoring primarily disrupts dialogue coherence, indicating that it imposes a necessary Global Constraint within the latent space that effectively mitigates entropy increase and objective divergence during long-text generation. While Template Retrieval exhibits a marginal impact on structure, it significantly optimizes the distributional priors of domain terminology via a Contextual Priming mechanism. Ultimately, the full model unifies macro planning, micro verification, and semantic grounding.

\section{Conclusion}
\label{sec:conclusions}
LLMs often underperform in CBT by defaulting to passive acknowledgments, asking few—often generic—questions, and failing to scaffold clients’ reasoning. We present the SIF, a controller that decouples when to intervene 
from what to ask 
, enabling interpretable control without end-to-end retraining; alongside, we curate a high-quality Socratic-QA dataset to sharpen specificity and reduce boilerplate. Empirically, SIF yields higher guidance accuracy, tighter domain relevance, and more purposeful question sequences than strong LLM baselines, improving alignment with CBT objectives and the progression from problem framing to actionable insights.

\section*{Limitations}
Despite encouraging results, our study has several limitations. SIF is trained primarily on Chinese-language counseling data, which may limit generalizability across cultures and languages. The current evaluation relies on simulated dialogues, and prospective validation in real-world clinical settings is still needed. The long-term effects of proactive questioning in ongoing therapeutic relationships also remain underexplored. Furthermore, assessments were conducted by one senior counselor and eight graduate students, which may introduce sampling and rater bias. Future work will expand to a broader, more diverse cohort of psychologists to enable a more comprehensive assessment.

\section*{Ethical Statement}
We integrated ethical safeguards at every stage of the study. All datasets were either publicly available or used under explicit authorization. We also verified that Sucratic-QA contains no personally identifying information. All data were anonymized to protect privacy and confidentiality in line with ethical standards, and informed consent was obtained for all relevant datasets.

SIF is intended to augment human counselors rather than replace them. Although it can meaningfully enhance proactive questioning in therapeutic dialogue, it does not substitute the professional judgment and expertise of licensed clinicians. Its role is decision support by surfacing timely strategies for consideration, while a licensed therapist retains oversight and ultimate responsibility for care. We are committed to deployments that meet the highest ethical standards, safeguard client wellbeing, and uphold the professional boundaries of psychological practice.

\textbf{Annotator Compensation.} \textit{During data collection, each annotator spent approximately 2 minutes per sample. We paid US\$0.319 per sample, corresponding to an effective hourly wage of ~US\$9.57 (~30 samples/hour), which exceeds the current U.S. federal minimum wage of US\$7.25/hour.}

\bibliography{custom}

\appendix
\label{sec:appendix}

\section{Psy-Insight Strategy}
\label{sec:Psy-Insight strategy}

\begin{figure}[th]
    \centering
    \includegraphics[width=\linewidth]{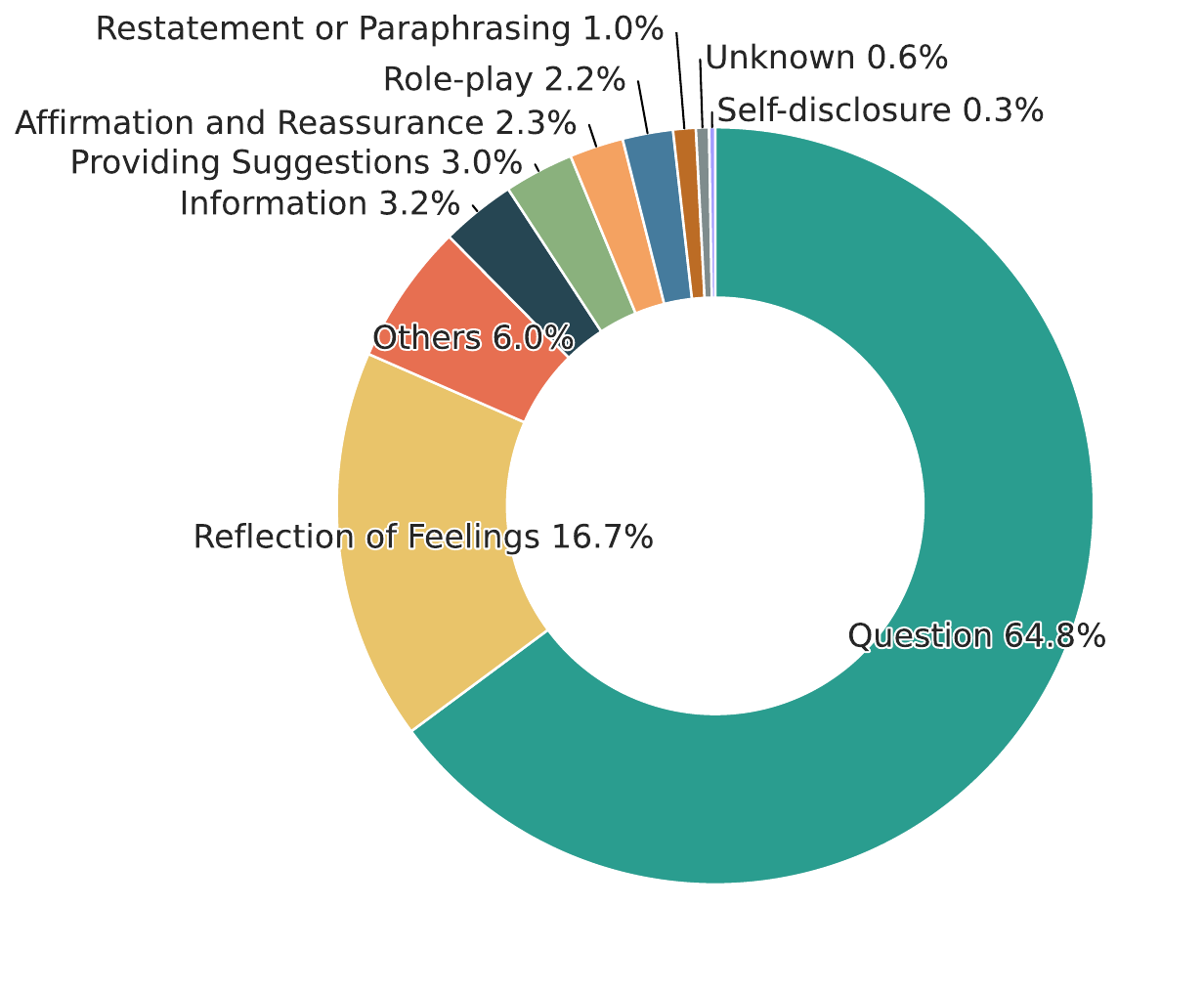}
    \caption{Distribution statistics of Psy-Insight strategy.}
    \label{fig:strategy_distribution_pie}
\end{figure}

As shown in Figure~\ref{fig:strategy_distribution_pie}, Psy-Insight is an interpretable multi-turn counseling dataset, originally released with both English and Chinese annotations; in this study, however, we only utilize the Chinese subset (CN). The Chinese portion comprises 431 sessions with 5,776 turns. Each session/turn is annotated with multi-task labels, including psychotherapy method, strategy, emotion, and topic, as well as explanatory notes. The strategy distribution exhibits a long-tail pattern, dominated by the Question strategy (64.8\%), while most other strategies appear with much lower frequency.

\section{Distribution of Samples}
\label{sec: Distribution of Samples}
\begin{figure}[th]
    \centering
    \includegraphics[width=\linewidth]{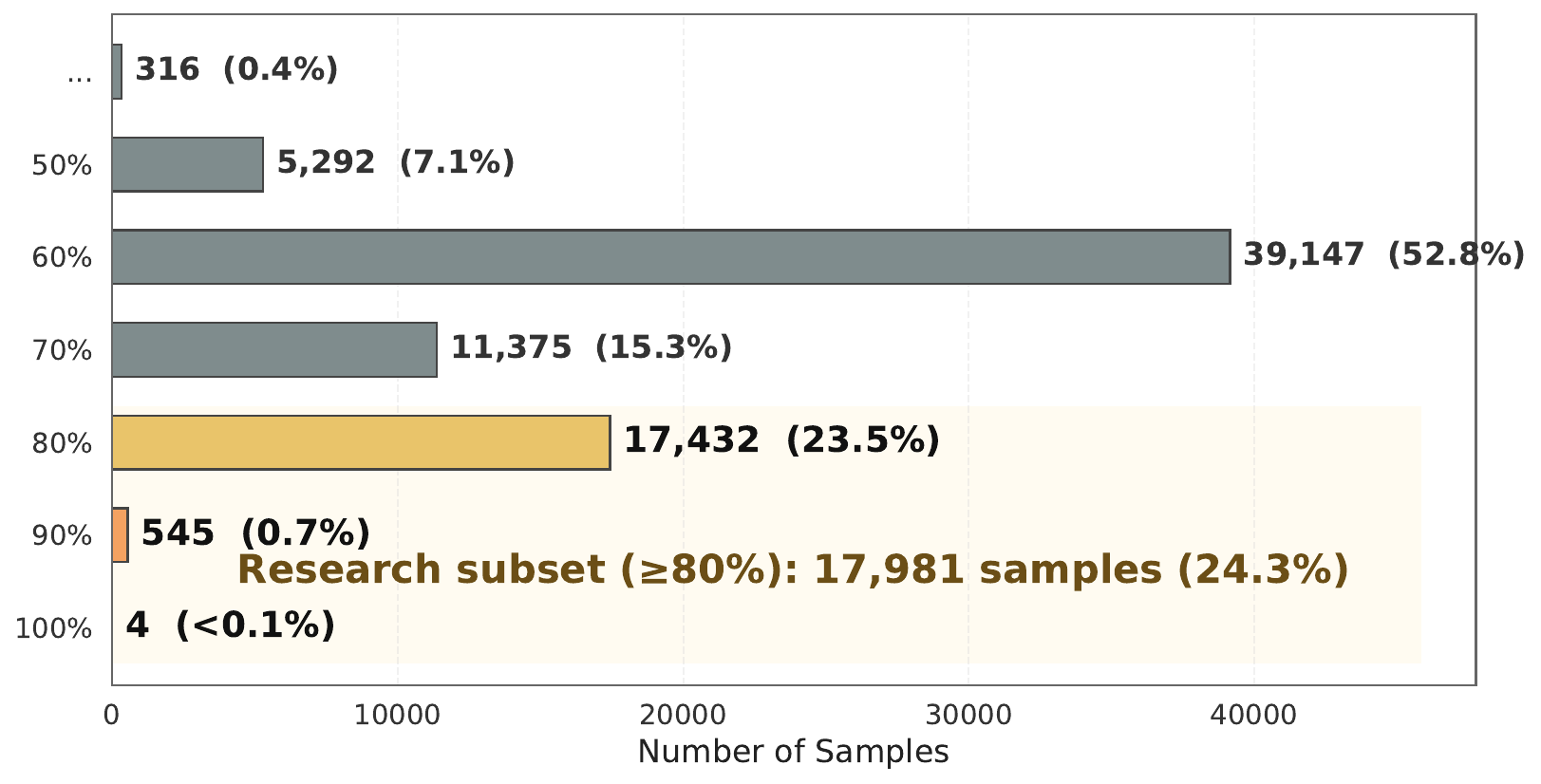}
    \caption{Distribution of multi-dimensional quality scores.}
    \label{fig:data_filter}
\end{figure}

As shown in Figure~\ref{fig:data_filter}, using multi-turn emotional dialogues from EmoLLM~\cite{53yang2024emollm}, we construct Socratic-QA by extracting counselor-centered contexts, generating context-aware questions via Socratic templates, and performing quality-based filtering to ensure robust generalization and prevent data leakage.

\section{Contrastive Preference Mechanism }
Algorithm~\ref{alg:quality_filtering} outlines the model first generates two distinct candidates: a direct inquiry ($q$) and a transition-enhanced variant ($Q$). To assess their potential impact, the algorithm constructs hypothetical future conversation states ($Conv_{1,2}$) by appending each candidate to the existing history. These states are then evaluated via a weighted expert scoring system, which aggregates performance across seven distinct metrics. Ultimately, the algorithm compares the cumulative scores to select the candidate yielding the higher utility ($C$) while discarding the inferior option ($R$), thereby ensuring the final output aligns with the system's multi-dimensional alignment goals.

\begin{algorithm}[tb]
\caption{Quality Filtering}
\label{alg:quality_filtering}
\textbf{Input:} Conversation $D$\\
\textbf{Parameter:} Emotional conversation model $ECM$, Format prompt for question $F_1$, Format prompt for transition $F_2$, Emotional parameter $w_i$\\
% \textbf{Parameter:} $w_{\text{empathy}}$, $w_{\text{guidance}}$ \\
\textbf{Output:} Chosen question $C$, rejected question $R$
\begin{algorithmic}[1]
    \STATE $prompt_1 = F_1(D)$
    \STATE $q = ECM(prompt_1)$
    \STATE $prompt_2 = F_2(D, q)$
    \STATE $Q = ECM(prompt_2)$
    \STATE $Conv_1 = concat(D, q)$
    \STATE $Conv_2 = concat(D, Q)$
    \STATE $Score_1 = \sum\nolimits_{i=1}^{7} w_i \cdot Expert_i(Conv_1, D)$
    \STATE $Score_2 = \sum\nolimits_{i=1}^{7} w_i \cdot Expert_i(Conv_2, D)$
    \IF{$score_1 > score_2$}
        \STATE $ C = q$, $ R = Q$
    \ELSIF{$score_1 < score_2$}
        \STATE $ C = Q$, $ R = q$
    \ELSE
        \STATE $ C = q$, $ R = None$
    \ENDIF
    \STATE \textbf{return} $(C, R)$
\end{algorithmic}
\end{algorithm}

\section{Socratic Templates}
\label{sec:TR model}
The Socratic prompting templates are derived through an expansion of the Psy-Insight dataset. We first leverage the dataset’s strategy annotations to guide the generation of candidate prompts as despite in Figure~\ref{fig:prompt-cards}. These candidates are then manually reviewed and filtered to ensure fidelity and therapeutic appropriateness. The curated collection serves as the training resource for the Template Retrieval (TR) module.

\section{Comparison Between Various Models}
\begin{figure}[th]
    \centering
    \includegraphics[width=\linewidth]{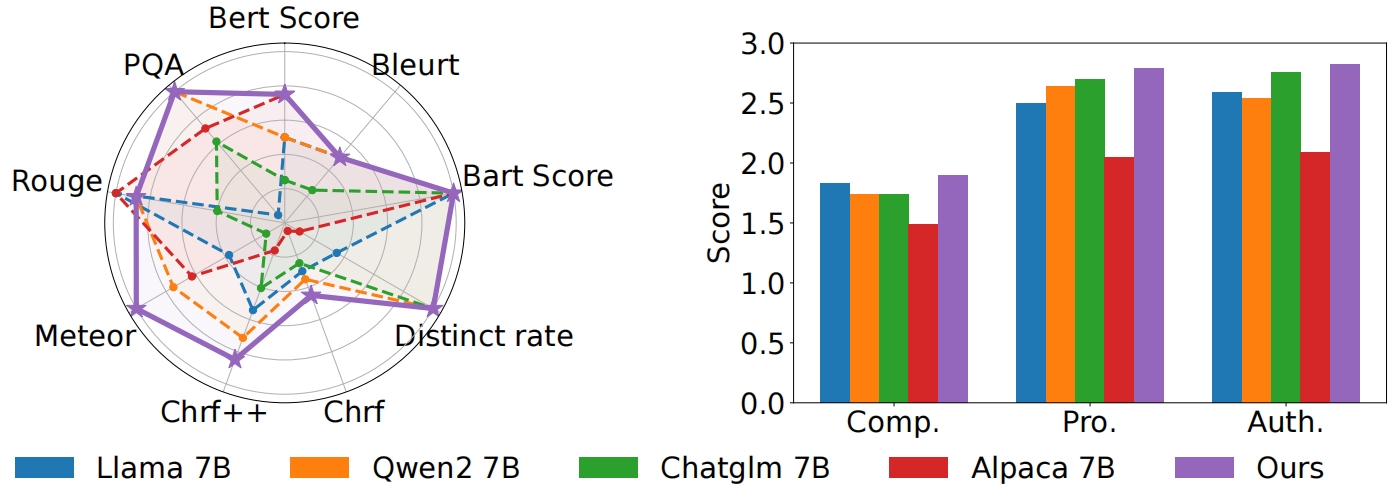}
    \caption{Comparison of general and professional capabilities across models. The left chart shows general language quality via automated metrics, while the right highlights domain-specific performance across four dimensions.}
    \label{fig:Strategy}
\end{figure}

\begin{figure*}[th]
  \centering
  \begin{promptbox}{Prompt}
\textbf{Instruction.} You are a \textbf{CBT therapist \& dialogue analyst}. Judge only the \textbf{last utterance} (its strategy is \texttt{\{example["strategy"]\}}) and output \textbf{JSON only}.

\textbf{Five-Step Guide (triggers \& goals):}
\begin{description}
  \item[1. Definition] Trigger: absolutes (``always'', ``completely''); 
  \\ Goal: baseline; Strategy: questioning, feeling reflection.
  \item[2. Elenchus] Trigger: cognitive distortions (all-or-nothing, catastrophizing); 
  \\ Goal: core belief; Strategy: questioning, paraphrase.
  \item[3. Maieutics] Trigger: uncertainty (``maybe'', ``possibly''); 
  \\ Goal: alternatives; Strategy: role-play, exploratory prompts.
  \item[4. Dialectics] Trigger: contradictory evidence; 
  \\ Goal: cognitive conflict; Strategy: advice, questioning.
  \item[5. Counterfactual] Trigger: conditionals (``If ... what then?''); 
  \\ Goal: reality testing; Strategy: info giving, self-disclosure.
\end{description}
\textbf{Priority:} use only the last utterance; 

\textbf{Required JSON:}
\\ 
\{"SocraticMethod":"%
Definition Method\textbar Elenchus\textbar Maieutics\textbar Dialectics\textbar Counterfactual Reasoning"\}

  \end{promptbox}
  \caption{Prompt cards for system and user instructions.}
  \label{fig:prompt-cards}
\end{figure*}

\section{Human Evaluation Process}
\label{sec:Human Evaluation Process}
Existing psychological LLMs can already generate proactive questions such as seeking subjective or objective information when given proper prompts, including GPT-4 acting as a counselor. Thus, the key issue for SIF is not whether proactive strategies appear, but whether they are initiated at the right time and whether they enhance dialogue quality and therapeutic outcomes. Since passive responses are generally adequate, our evaluation emphasizes the effectiveness of proactive recommendations. To this end, we use four expert-informed metrics for assessing proactive strategies in psychological conversations.

\paragraph{Strategy Comprehensivenes (0–--2 points).}
This dimension evaluates whether the reply accurately reflects the client’s basic information and core psychological issues. High scores require covering key elements such as main concerns, context, duration, severity, and functional impact, with evidence drawn from the client’s own statements. Partial or vague reflection lowers the score, while misinterpretation or disconnection from the dialogue history results in the lowest score.

\paragraph{Professionalism (0–--3 points).}
This dimension assesses the counselor’s professional judgment and techniques. High scores reflect clear clinical reasoning, appropriate psychological methods, professional guiding language, and practical help for the client. Following the standard counseling sequence (intake, diagnosis, intervention, consolidation) and presenting techniques in concrete, step-by-step detail can yield maximum points. Generic advice, disorganized processes, or superficial technique use reduce the score.

\paragraph{Authenticity (0–--3 points).}
This dimension examines whether the reply is consistent with multi-turn dialogue history, demonstrates empathy and understanding, and resembles a realistic counseling exchange. High scores require concise, empathic, and contextually relevant responses that avoid overpromising, lecturing, or causing discomfort. Irrelevant, mechanical, or misleading content lowers the score, as does excessive length inconsistent with real counseling interactions.

\paragraph{Ethical Safety (0–--1 point).}
This dimension ensures privacy protection and respect for the client. High scores demand strict confidentiality, avoidance of disclosing or requesting irrelevant sensitive details, and a respectful tone toward the client’s feelings and autonomy. Any privacy risks, disrespectful comments, or boundary violations result in a zero score.

\subsection{Psychological Expert Evaluation Process}
Our evaluation team included eight graduate students and one senior counselor. All received standardized training, with the senior counselor supervising scoring and resolving disputes. Before participation, evaluators signed consent agreements (Figure 9). They selected topics, reviewed supporting datasets to construct realistic client profiles and opening statements, and then role played as clients in sessions of at least ten turns with each model under two conditions: SIF enhanced and baseline. Ratings were logged immediately after each turn, and an independent processor aggregated scores to compute per-metric averages; substantial discrepancies were adjudicated by the senior counselor.

\section{Implementation}
\label{sec:Implementation}
\paragraph{Implementation.} The LPP module is trained on strategy  annotations from the Psy-Insight dataset using AdamW (learning rate=1e-5). For SIF, the CG component is instantiated with Qwen2.5-7B-Instruct~\cite{55bai2023qwen}, fine-tuned on the Socratic-QA training set for three epochs with LoRA. All models use consistent decoding parameters: temperature=0.5, top\_p=0.75, top\_k=20. Multi-turn dialogues are simulated using Deepseek-V3 as a client that generates seeker utterances based on predefined emotional profiles.

\section{Performance}
\label{sec:Additional}
Automatic metrics in Table~\ref{tab:tab_stage1_stage2} are consistent with these findings, showing better accuracy, precision, and F1 for both anchoring and retrieval tasks.
%-------------------------------------------------------------------
\begin{table}[ht]
\centering
% \vspace{0.5em}
\scalebox{0.8}{
\begin{tabular}{lcccccc}
\toprule
 & \multicolumn{3}{c}{\textit{Strategy Anchor}} & \multicolumn{3}{c}{\textit{Template Retrieval}} \\
\cmidrule(lr){2-4} \cmidrule(lr){5-7}
\textbf{Method} & \textbf{Acc} $\uparrow$ & \textbf{Pre} $\uparrow$ & \textbf{F1} $\uparrow$ & \textbf{Acc} $\uparrow$ & \textbf{Pre} $\uparrow$ & \textbf{F1} $\uparrow$ \\
\midrule
BERT & 0.716 & 0.595 & 0.636 & 0.463 & 0.489 & 0.472 \\
T5 & 0.661 & 0.615 & 0.629 & 0.459 & 0.483 & 0.468 \\
Ours & \textbf{0.721} & \textbf{0.637} & \textbf{0.669} & \textbf{0.485} & \textbf{0.511} & \textbf{0.494} \\
\bottomrule
\end{tabular}
}
\caption{Performance comparison of models on automatic evaluation metrics.}
\label{tab:tab_stage1_stage2} 
% \vspace{-1em}
\end{table}
%-------------------------------------------------------------------

\subsection{Policy Anchoring Performance}
As shown in Table~\ref{tab:tab_stage1_stage2}, the classifier identifies counseling strategies significantly better than chance, despite the inherent difficulty of the task. Qwen achieves the highest performance with 72.1\% accuracy and an F1 score of 0.669, outperforming BERT by +0.5 percentage points in accuracy and +3.3 in F1. In contrast, T5 lags by 6 points in accuracy, indicating that span-generation pre-training is less effective for conversation-policy classification than instruction-tuned objectives.

\subsection{Template Retrieval Performance}
Using an identical encoder in the rule-based retriever exposes TR as the dominant bottleneck. As shown in Table~\ref{tab:tab_stage1_stage2}, Qwen achieves only 48.5\% accuracy and an F1 score of 0.494. In contrast, T5 variants reduce the gap to 2.6 percentage points, suggesting that retrieval performance is now limited more by dataset constraints than by encoder capacity.

\section{Case Study Details}
\label{sec:Case Study Details}
We demonstrate the effectiveness of SIF through a representative case study, with the original dialogue presented in Figure~\ref{fig:An example of the effect of the SIF application (Chinese version)} and its English translation in Figure~\ref{fig:An example of the effect of the SIF application (English version)}.

The primary limitation of the PsyLLM model lies in its overreliance on direct advice and reassurance, with minimal attention to the client’s underlying thoughts and emotions. Counselors following this approach seldom employ open-ended questioning; instead, they prescribe responses such as “you can say this” or “you should try self-affirmation.” This directive style encourages passive adoption of coping strategies rather than active exploration of root causes. Consequently, clients may acquire temporary techniques but fail to internalize the psychological mechanisms driving their behavior, thereby hindering deeper and more enduring change.

In contrast, the Socratic Inquiry Framework (SIF) prioritizes the cultivation of self-awareness. Through iterative exploratory questions (e.g., “What does it mean to refuse others?” or “What would your ideal state look like?”), SIF guides clients to identify self-imposed constraints and clarify authentic needs, progressively strengthening autonomy and intrinsic motivation. While the Psy model provides concrete but surface-level solutions, it rarely facilitates the development of new cognition or sustainable transformation. In essence, the Psy model emphasizes giving answers, whereas SIF emphasizes eliciting reflection, aligning more closely with professional standards in psychological counseling.

\begin{figure*}
    \centering
    \includegraphics[width=\linewidth]{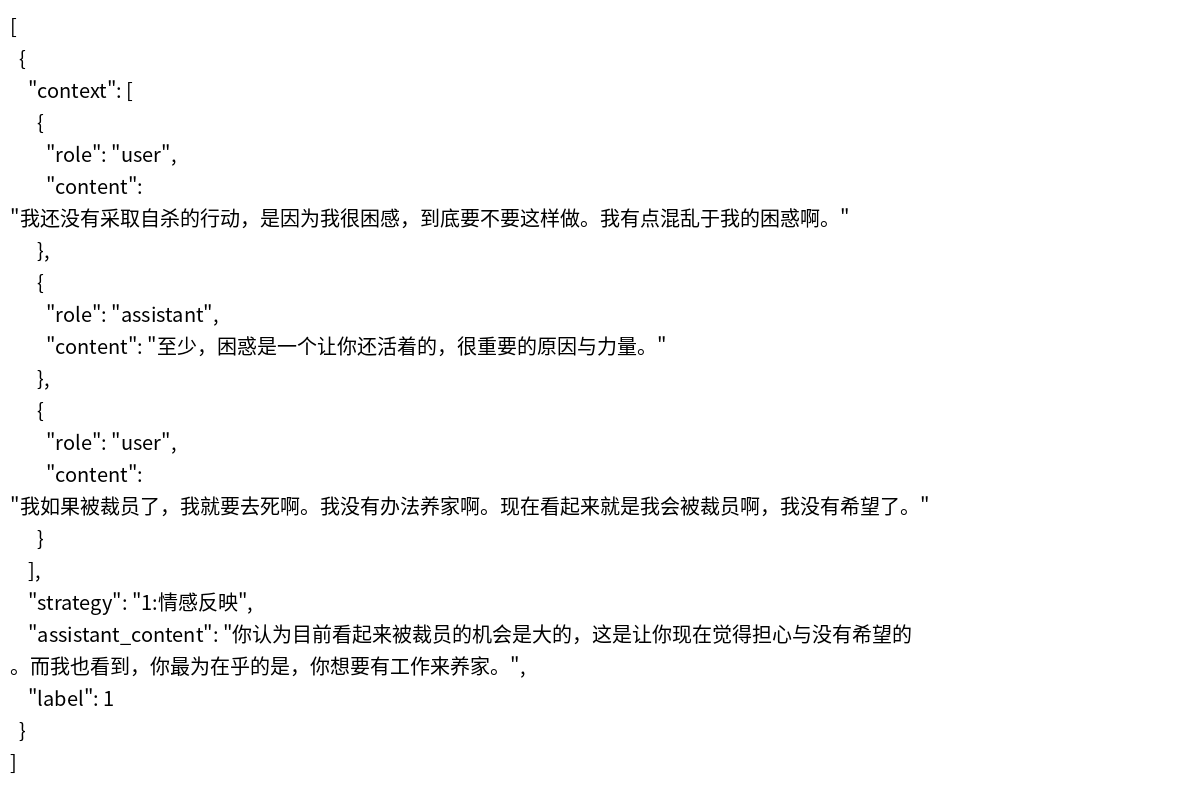}
    \caption{Example of Strategy (Chinese version).}
    \label{fig:Example of Strategy (Chinese version)}
\end{figure*}

\begin{figure*}
    \centering
    \includegraphics[width=\linewidth]{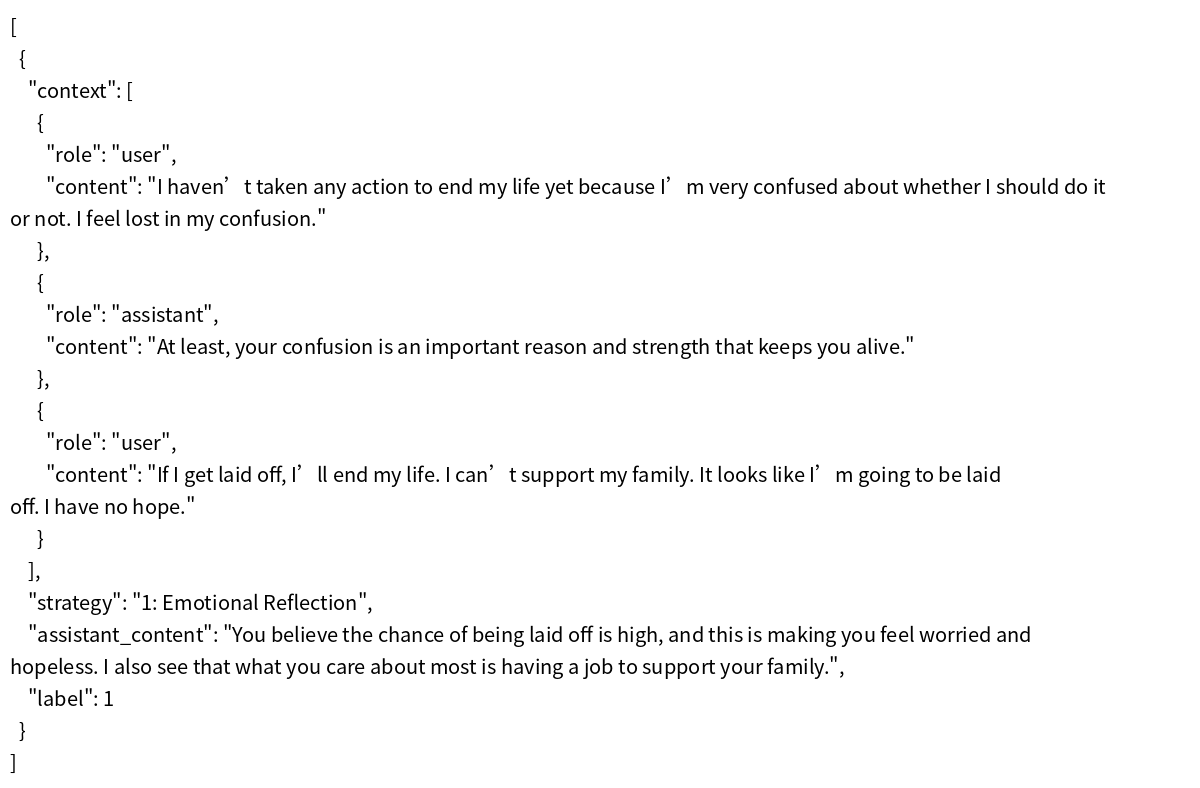}
    \caption{Example of Strategy (English version).}
    \label{fig:Example of Strategy (English version)}
\end{figure*}

\begin{figure*}
    \centering
    \includegraphics[width=\linewidth]{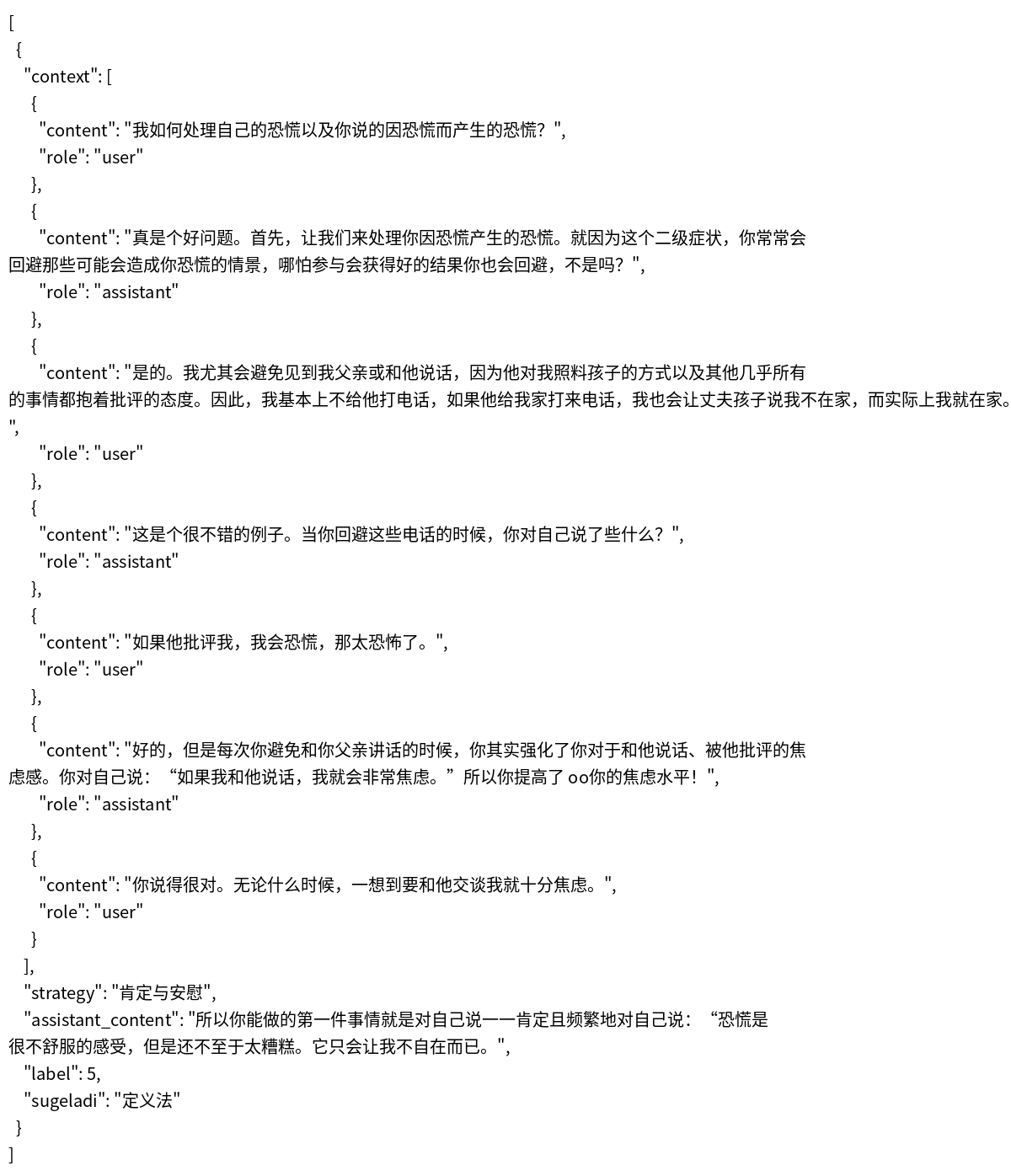}
    \caption{Example of Socratic Template (Chinese version).}
    \label{fig:Example of Socratic Template (Chinese version)}
\end{figure*}

\begin{figure*}
    \centering
    \includegraphics[width=\linewidth]{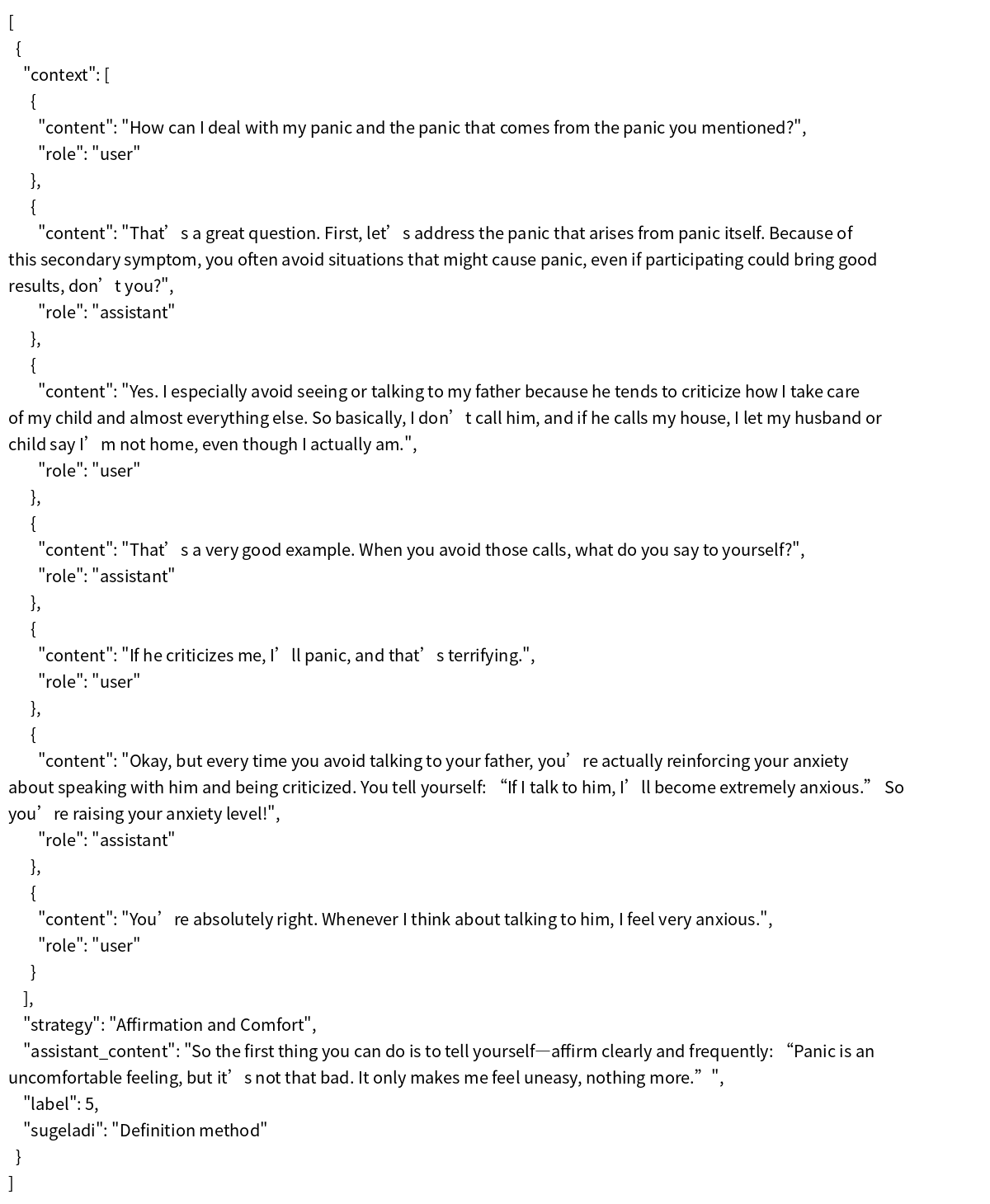}
    \caption{Example of Socratic Template (English version).}
    \label{fig:Example of Socratic Template (English version)}
\end{figure*}

\begin{figure*}
    \centering
    \includegraphics[width=\linewidth]{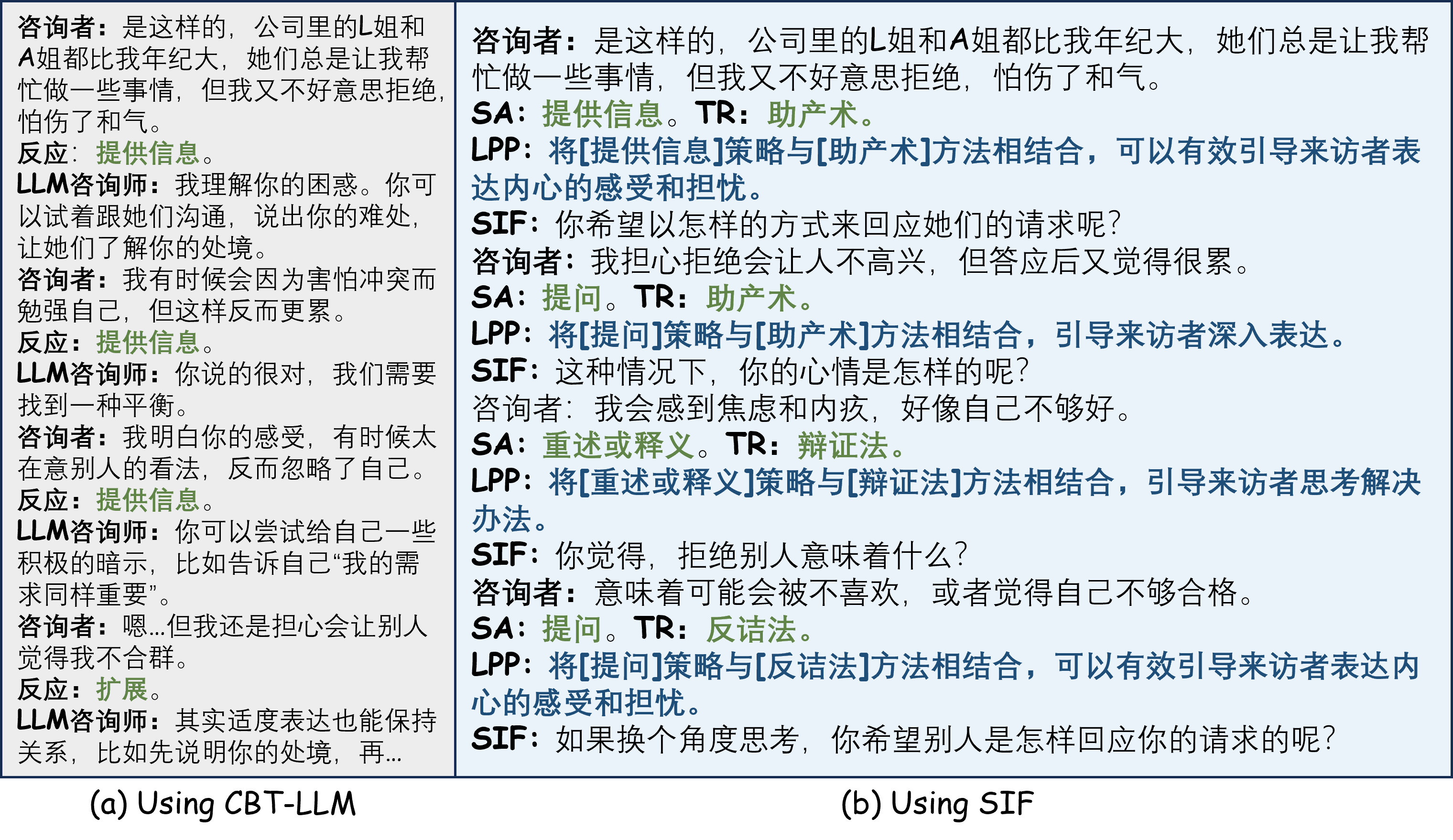}
    \caption{An example of the effect of the SIF application (Chinese version).}
    \label{fig:An example of the effect of the SIF application (Chinese version)}
\end{figure*}

\begin{figure*}
    \centering
    \includegraphics[width=\linewidth]{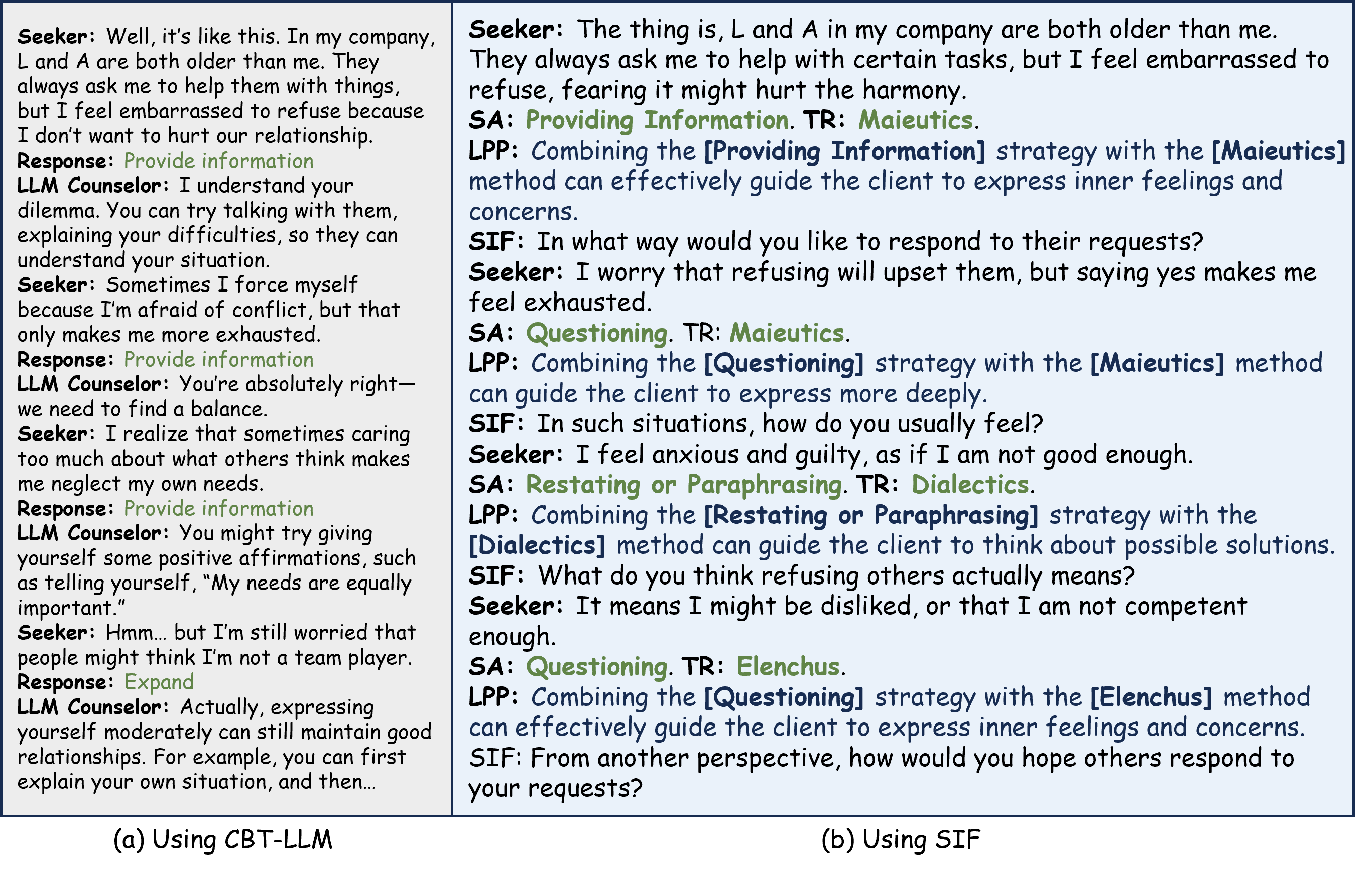}
    \caption{An example of the effect of the SIF application (English version).}
    \label{fig:An example of the effect of the SIF application (English version)}
\end{figure*}

\begin{figure*}
    \centering
    \includegraphics[width=\linewidth]{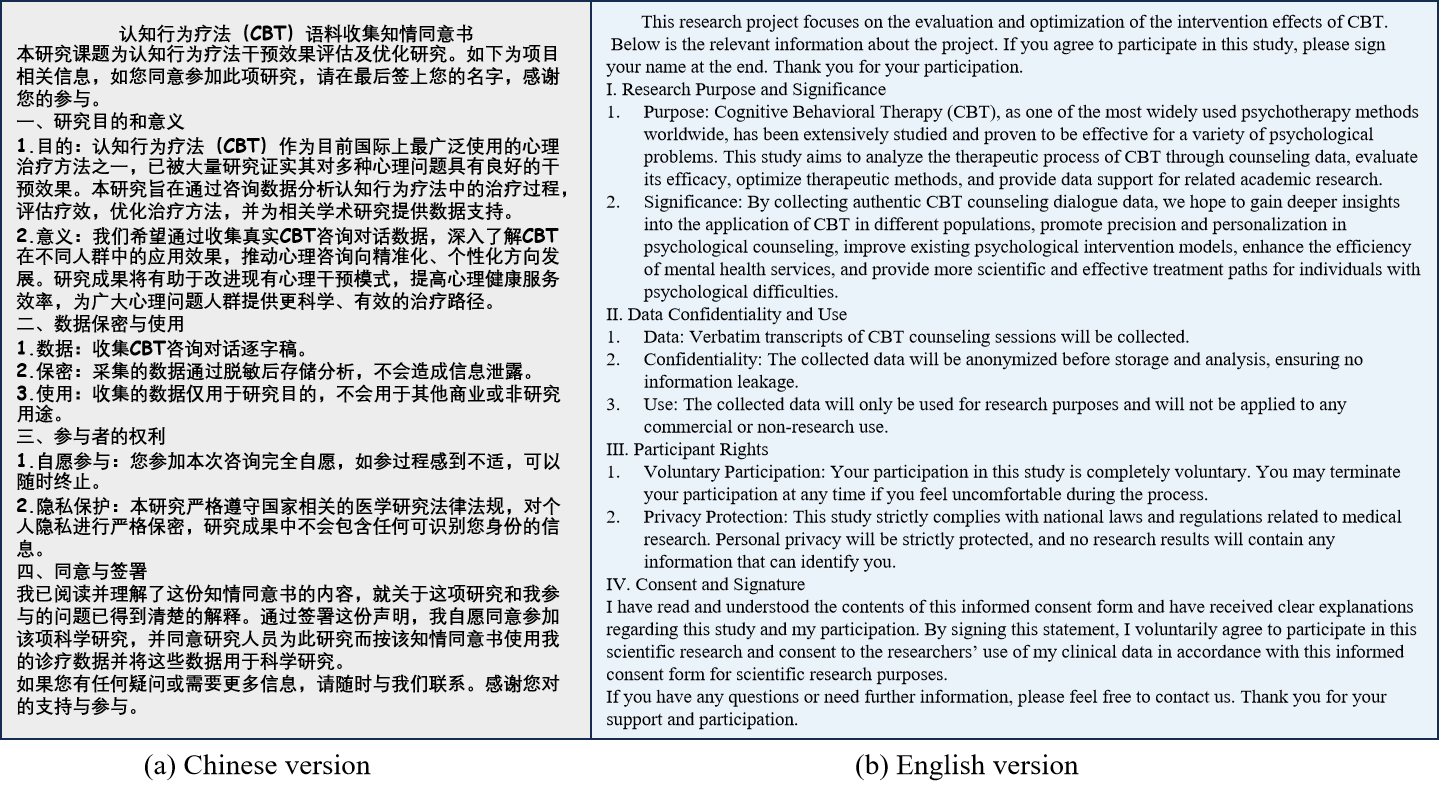}
    \caption{Content of the Participant Informed Consent.}
    \label{fig:Content of the Participant Informed Consent}
\end{figure*}

\end{document}